\documentclass[standards,article,submit,pdftex,moreauthors]{Definitions/mdpi} 
\firstpage{1} 
\pubvolume{1}
\issuenum{1}
\articlenumber{0}
\pubyear{2025}
\copyrightyear{2025}
\datereceived{ } 
\daterevised{ } 
\dateaccepted{ } 
\datepublished{ } 
\hreflink{https://doi.org/} 

\usepackage{dsfont}

\Title{Automated Road Distress Detection Using Vision Transformers and Generative Adversarial Networks}

\TitleCitation{Automated Road Distress Detection Using Vision Transformers and Generative Adversarial Networks}

\Author{Cesar Portocarrero Rodriguez $^{1,\ddagger}$*\orcidA{}, Laura Vandeweyen $^{1,\ddagger}$ and Yosuke Yamamoto $^{1,\ddagger}$}

\AuthorNames{Cesar Portocarrero Rodriguez, Laura Vanderweyen and Yosuke Yamamoto}

\isAPAStyle{%
       \AuthorCitation{Portocarrero, C., Lastname, F., \& Lastname, F.}
         }{%
        \isChicagoStyle{%
        \AuthorCitation{Portocarrero Rodriguez, Cesar, Firstname Lastname, and Firstname Lastname.}
        }{
        \AuthorCitation{Portocarrero Rodriguez, C.; Vanderweyen, L.; Yamamoto, Y.}
        }
}

\address[1]{%
$^{1}$ \quad Stanford University, Department of Civil and Environmental Engineering\\
}

\corres{Correspondence: cportoca@stanford.edu}

\secondnote{These authors contributed equally to this work.}

\abstract{The American Society of Civil Engineers has graded America’s infrastructure condition as a C, with the road system receiving a dismal D. Roads are vital to regional economic viability, yet their management, maintenance, and repair processes remain inefficient, relying on outdated manual or laser-based inspection methods that are both costly and time-consuming. With the increasing availability of real-time visual data from autonomous vehicles, there is a significant opportunity to harness computer vision (CV) for advanced road monitoring, offering insights to guide infrastructure rehabilitation efforts. This project explores the application of state-of-the-art CV techniques for road distress segmentation. It begins by evaluating the use of synthetic data generated through Generative Adversarial Networks (GANs) to assess the effectiveness of such data for model training. The study then applies Convolutional Neural Networks (CNNs) for road distress segmentation and subsequently implements the vision transformer model MaskFormer, identifying areas for potential improvement in road monitoring. Results demonstrate that GANs are a viable method for generating synthetic data, significantly enhancing the performance of CV models. Further, MaskFormer shows promise in this task, as it outperforms the CNN model in two metrics: mAP50 and IoU.}


\begin{document}

\section{Introduction}

In 2025, the American Society of Civil Engineers attributed a score of C to America's infrastructure conditions \cite{ASCE_report_card}. The solution to this problem, as well as its safety and cost implications, requires leveraging technology to improve management programs and prioritize maintenance operations, asset replacement, and investment. Technological initiatives, however, cannot be general; they must be tailored to the type of infrastructure being studied. This specific research paper focuses on road infrastructure. The road system, in particular, received a score of D, as 40\% of the system today is in poor or mediocre conditions. Each year, there are 11K deaths and \$3B in damages due to poor road conditions. Additionally, it is 14 times more expensive for public agencies to reconstruct roads than it is to perform preventive maintenance over its life cycle \cite{FHA_PublicRoads}.

Despite the importance of infrastructure maintenance to sustain the economy and provide a great return on taxpayer money, inspection processes remain archaic. Public institutions in charge of the transportation system often have to execute manual road inspections, whereby an engineer drives the road and stops at relevant locations to document a particular defect. As expected, this process is considerably time-consuming and cost-inefficient due to the high number of manhours required. Further, the repetitive nature of the job means that human workers grow rapidly weary during the process, leading to a decrease in accuracy and an increase in misannotations when documenting cracks, potholes, and other sources of road distress.

To improve the performance in predicting road deterioration, state government agencies such as CalTrans also rely on additional tools such as laser scanners \cite{soilan2019review}. Laser inspections, however, are usually performed only once per year since they require a dedicated driver, an engineer, and a specialized vehicle heavily equipped with machinery. The data collected through this laser detection process allows the International Road Roughness Index (IRR) to be derived at N discrete points per mile of lane-road. In addition to its cumbersome deployment, the performance of laser technology for road distress detection leaves much to be desired. Public utility transportation engineers complain that the laser data sometimes does not differentiate between open cracks and cracks that have already been sealed.

In response, recent years have seen the development of research seeking to apply computer vision solutions to the problem of road distress detection \cite{Alpher02, Alpher03, Alpher05, Alpher06}. The development of semantic segmentation algorithms for road assessments has the potential to significantly improve maintenance programs of transportation systems. In this project, we used road images form the National Department of Transport Infrastructure (NDTI) to explore the application of SOTA techniques such as CNNs, GANs, and vision transformers to output road distress categories, namely cracks, potholes, damaged markings, and guardrails. 
\section{Related work}

\subsection{Generative adversarial networks}
GANs (Generative Adversarial Networks) are a popular deep learning framework consisting of two neural networks: a generator and a discriminator that compete against each other. It was introduced by Ian Goodfellow in 2014  \cite{goodfellow2014generative}. GANs have been extensively researched and applied to various domains, including image synthesis, text generation, and video generation. They have shown remarkable success in generating realistic and diverse samples by leveraging an adversarial learning process \cite{DamageDecectionGANs}. However, GAN training can be challenging, often suffering from mode collapse and instability issues. Many techniques have been proposed to address these challenges, such as Wasserstein GANs, progressive growing, and spectral normalization. GANs continue to be an active area of research, with ongoing efforts to improve training stability, sample quality, and application versatility. It is worth noting that GANs have not been used to upsample road imagery data to date.

\subsection{Vision transformers}
Transformers were first introduced for machine translation in the domain of natural language processing. They are characterized by their encoder-decoder structure without recurrence and the adoption of self-attention \cite{TransformersOrig}. In the field of computer vision, the Vision Transformer (ViT) was introduced as an adaptation of the original architecture for image classification. Images are split into patches and then the ViT can extract global features throughout the layers. Results showed that the ViT was able to surpass the performance of SOTA CNN architectures \cite{dosovitskiy2021image}.

In the context of road infrastructure, the first experiment with vision transformers was carried out for road surface condition detection. Although the performance of this model is remarkably high, with an F1-score of 0.9671, it is worth noting that it is limited to a classification problem that only contemplates 3 classes: heavy rain, light rain, and no rain. Further, the performance benefits of an artificial spatial self-attention network to model the relationship between adjacent images \cite{rain}. 

Regarding distress detection on road infrastructure, the literature is limited to binary classification tasks involving two classes: cracked and healthy asphalt. A ViT-based framework proposed by Asadi et al. achieved an IoU of 61\%, an increase of 3.8\% compared to the best-performing CNN-based model \cite{transcrack1}. It is important to note that this model was trained using online datasets that are artificially centered on cracks, minimizing the noise. A second experience was carried out by Xiao et al. using hybrid-window attentive vision transformers \cite{Crackformer}. This framework, named CrackFormer, aims to extract feature semantics locally and globally with dense and sparse windows, respectively. The results show that CrackFormer performs significantly well, with an F1-score of 0.9364. However, similar to the previous research, the data is based on internet datasets that do not capture a realistic view of road images taken by a dedicated vehicle such as a car or a UAV. Moreover, these models are limited to image classification tasks, meaning that they are unable to capture the real shape and dimensions of road distress, factors that can be extracted via semantic segmentation and are important for material takeoff. Finally, these models ignore sources of road distress other than cracks, leading to the uncertainty of whether transformers will also perform well with potholes, marking damage, etc. 

\subsection{Convolutional neural networks}
Semantic segmentation of road conditions, such as cracks, potholes, and road markings is a challenging task due to the complex and varied nature of these conditions. To overcome these challenges, advanced computer vision techniques such as deep learning and convolutional neural networks have been applied in the literature, showing promising results in pavement distress detection. For example, the previously cited study uses an algorithm called MobileCrack to perform object classification for the maintenance of asphalt pavement surfaces, achieving an accuracy of 0.865 \cite{Alpher02}. Another study also uses data collected from UAVs to perform semantic segmentation for pavement defect detection using algorithms such as Faster R-CNN, YOLOv3, and YOLOv4, and achieves comparable accuracy results \cite{Alpher03}.  A Deep CNN detection model achieved an accuracy of 0.9375 but with a high risk of over-fitting given a data-set size smaller than 300 images \cite{Alpher05}.

Despite the inmense progress, however, there are still challenges in the current literature. For example, inaccuracies in annotation may affect the performance of the algorithm if the quality of the data is insufficient, an observation many researchers have noted on local datasets. Also, the robustness of the algorithm to the effects of weather and lighting conditions needs to be improved. In addition, more advanced deep learning architectures need to be developed to improve the accuracy of semantic segmentation, which is one of the directions for future research.

Further, there is limited research about road marking damage detection, in addition to a  scarcity of annotated data \cite{Alpher06}. Moreover, some road markings need to be repaired in the future, while others need to be erased. Currently, no algorithm has been developed to make that distinction.
\section{Materials and Methods}

The proposed methodology is split into three parts. First, a GAN was trained to augment the dataset with synthetic images. It is expected that these plausible and realistic road images will improve the performance of the algorithms compared to the SOTA which just relies on standard data augmentation techniques (random flips, rotations, etc.). Then, semantic segmentation was carried out on the augmented dataset using two distinct methodologies: YOLOv8 and MaskFormer. The former is a CNN-based model that serves as a benchmark to evaluate the performance of the latter, a visual transformer-based semantic segmentation algorithm. Since the SOTA has not examined the performance of visual transformers on semantic segmentation tasks in road infrastructure, the proposed methodology aims to address this objective.

\subsection{Synthetic data generation using GANs}
Unlike typical imagery used in computer vision projects, road datasets contain features such as pavement, cracks, and guardrails that are continuous in nature and highly imbalanced in their representation—for example, pavement is vastly overrepresented compared to cracks or guardrails. These unique characteristics call for tailored approaches in both the design of the GAN architecture and the learning algorithm.

\subsubsection{Building the Generator and Discriminator Models}
\begin{figure}[h]
  \centering
  \includegraphics[width=0.6\linewidth]{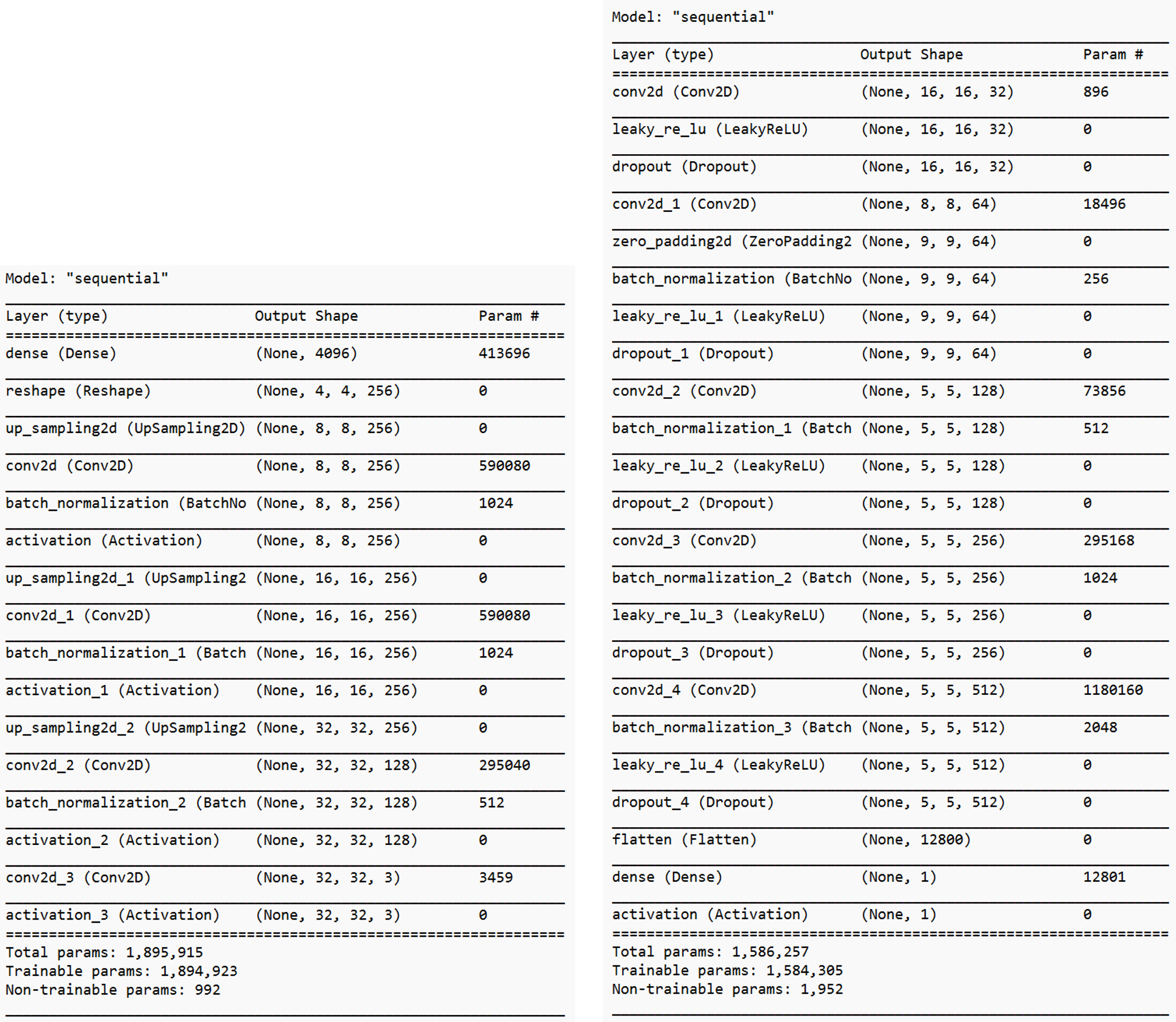}
   \caption{Architecture selected for the generator and discriminator models.}
   \label{fig:gendisc}
\end{figure}

Training a GAN model is accomplished by two networks, a generator and a discriminator that compete with each other. The Generator is responsible for generating new data. The input layer consists of a fully connected layer (Dense) with ReLU. A Reshape layer is added to reshape the output for further processing. UpSampling2D layers are used to increase the resolution of the image.
Conv2D layers are used for feature extraction. Additional UpSampling2D and Conv2D blocks can be added to further increase resolution and extract features. 
The discriminator is responsible for distinguishing between real and fake images. The input layer is a Conv2D layer that matches the shape of the specified image, reducing its size with a stride of 2. LeakyReLU is applied to introduce nonlinearity and allow the model to learn a wider range of features. A dropout layer is added to prevent overfitting by randomly disabling neurons during training. Multiple Conv2D and BatchNorm layers extract image features with a LeakyReLU activation for nonlinearity. Then, a flattening layer transforms the image data into a 1D vector. The output layer is a single fully connected layer (Dense) with Sigmoid, outputting a probability to determine if the generated image is real or fake. Figure~\ref{fig:gendisc} shows a summary of the architecture for both the Generator and Discriminator models.

\subsubsection{Defining Loss Functions and Optimizer}
Training the Discriminator requires the ability to distinguish the real images from the fake ones. For images generated by the Generator, the 0 label is the output for fake data, while the 1 label is assigned to real images. The loss function is then defined as binary cross-entropy loss. The discriminator loss function calculates the total loss by comparing the real and generated outputs. The generator loss is calculated by comparing the generated output with the expectation of being classified as real.

\subsection{YOLOv8 implementation}
YOLO (You Only Look Once) is a real-time object detection algorithm originally developed in 2015 by Joseph Redmon and Ali Farhadi. It employs a single neural network to simultaneously predict bounding boxes and class probabilities for objects within an image. Over time, multiple versions have been introduced, each improving upon the original. For this research, YOLOv8 was selected due to its stability and its frequent use in recent literature as a benchmark. This state-of-the-art model features a redesigned backbone network, an anchor-free detection head, and an updated loss function, offering high efficiency across a wide range of hardware platforms.\cite{DiveintoYOLOv8}.

\subsubsection{YOLOv8 backbone}
\begin{figure}[h]
  \centering
  \includegraphics[width=\linewidth]{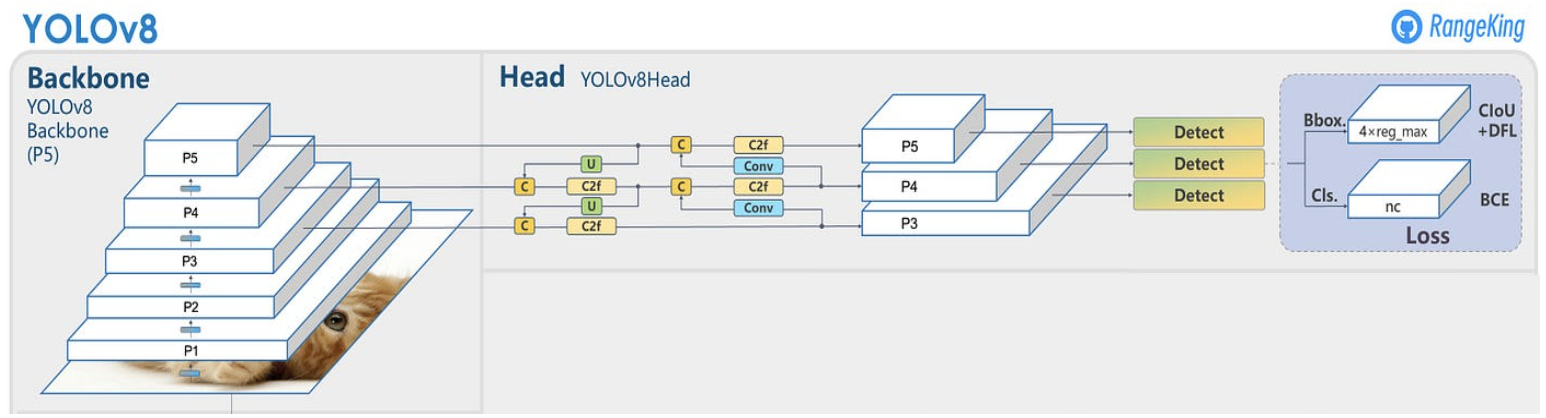}
   \caption{Diagram of YOLOv8 backbone architecture. \cite{Github_YOLOarch}}
   \label{fig:backbone}
\end{figure}

In YOLOv8's backbone, seen in Figure \ref{fig:backbone}, five primary convolutional layers (P1 to P5) use Kernel 3 and stride 2, differing from YOLOv5's P1 with Kernel 6. It replaces the Concentrated-Comprehensive Convolution (C3) blocks with Coarse to Fine (C2F) blocks after P2, 3, 4, and 5. Unlike C3's use of the last Bottleneck's output, C2f concatenates all Bottleneck outputs. In the neck, features are directly concatenated without matching channel dimensions, reducing parameter count and tensor size \cite{Roboflow_YOLOv8}.

\subsubsection{YOLOv8 anchor-free detection head}
\begin{figure}[h]
  \centering
  \includegraphics[width=\linewidth]{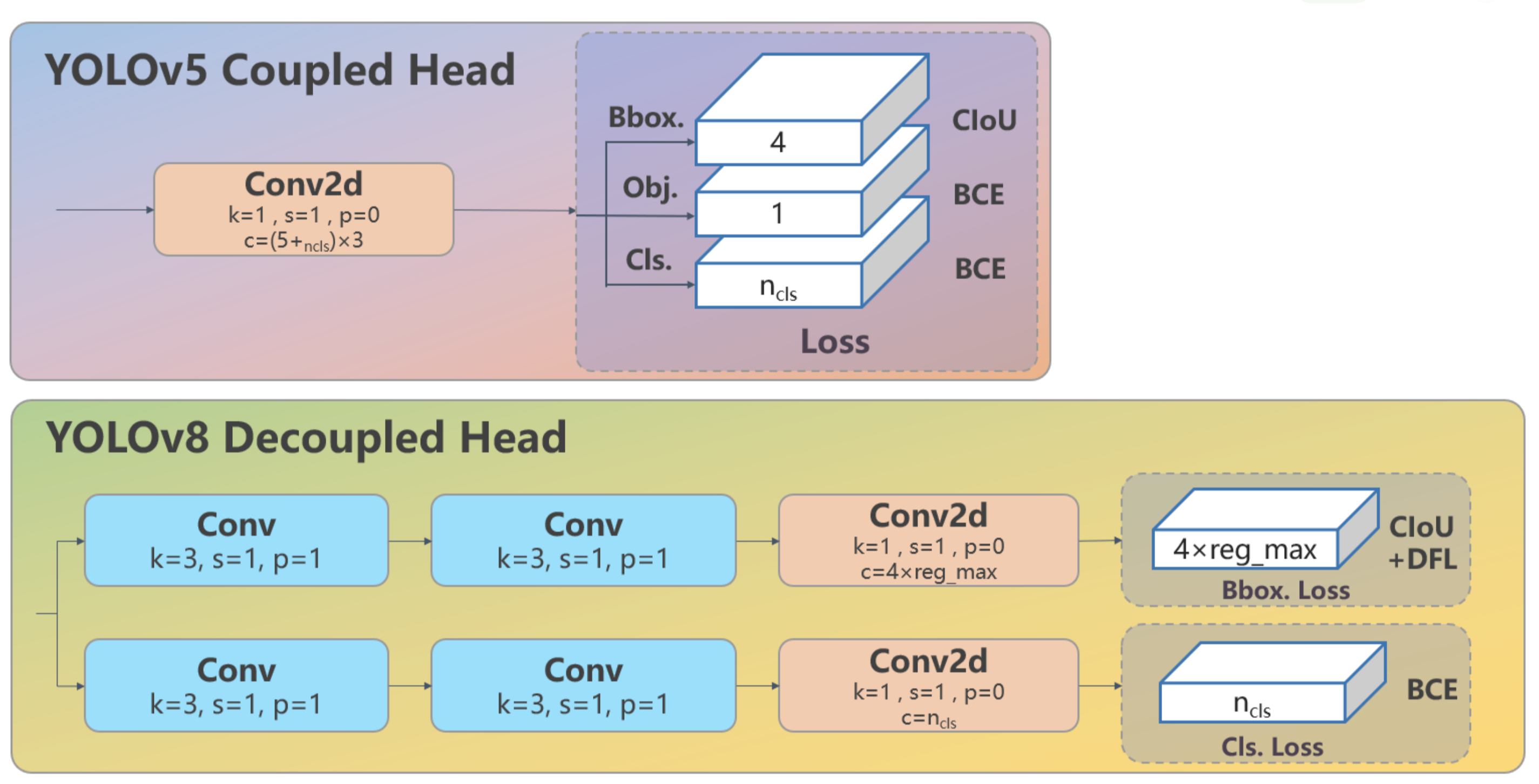}
   \caption{Detection head for Yolov8 \cite{Github_YOLOarch}}
   \label{fig:Yolov8arch}
\end{figure}

YOLOv8's head module, shown in Figure \ref{fig:Yolov8arch}, uses upsampled and re-concatenated outputs of C2f following P5-3, differing from YOLOv5's anchor-based structure. It uses an anchor-free design, generating and then refining bounding box predictions around objects. Anchor boxes reflect the benchmark's box layout, not the unique dataset's, and their absence reduces box predictions, speeding up the filtering process in Non-Maximum Suppression (NMS).

\subsubsection{YOLOv8 loss function}
In YOLOv8, the loss calculation uses a dynamic sample assignment strategy and a two-part loss calculation. It uses the Task-aligned One-stage Object Detection (TOOD) framework's TaskAlignedAssigner to select positive samples based on weighted classification and regression scores. Loss is computed for classification using Binary Cross Entropy (BCE) Loss, and for regression with Distribution Focal Loss and Complete IoU (CIoU) Loss, each weighted specifically, excluding the objectness loss seen in YOLOv5 (see Figure \ref{fig:Yolov8arch}).

\subsection{MaskFormer implementation}
Unlike typical per-pixel semantic segmentation paradigms, MaskFormer performs mask classification, in which it predicts a set of binary masks, each associated with a single class prediction \cite{MaskFormer}. 

\subsubsection{Mask classification formulation}
In a traditional per-pixel semantic segmentation model, the probability distribution over all possible $K$ categories for every pixel is calculated in an $H \times W$ image. Thus, the per-pixel cross-entropy loss is defined as shown in equation \ref{eq:loss_pixel}.

\begin{equation}
    \mathcal{L}_{pixel-cls}(y, y^{gt})=\sum^{HW}_{i=1}-\log p_i(y^{gt}_{i})
    \label{eq:loss_pixel}
\end{equation}

In contrast, mask classification uses binary masks defined as $\{m_i|m_i\in [0,1]^{H \times W} \}$ for each category on every image. Further, each region is associated as a whole (not per-pixel) with some probability distribution over each of the $K$ categories. Altogether, the total loss is a combination of cross-entropy classification loss and binary mask loss for each predicted segment as presented in equation \ref{eq:loss_mask}.

\begin{equation} 
    \mathcal{L}_{mask-cls}(z, z^{gt})=\sum^{N}_{j=1}[-\log p_{\sigma(j)}(c^{gt}_{j})+1_{c^{gt}_{j}\neq \emptyset}\mathcal{L}_{mask}(m_{\sigma(j)}, m^{gt}_j)]
    \label{eq:loss_mask}
\end{equation}

\subsubsection{MaskFormer architecture}
\begin{figure}[h]
  \centering
  \includegraphics[width=\linewidth]{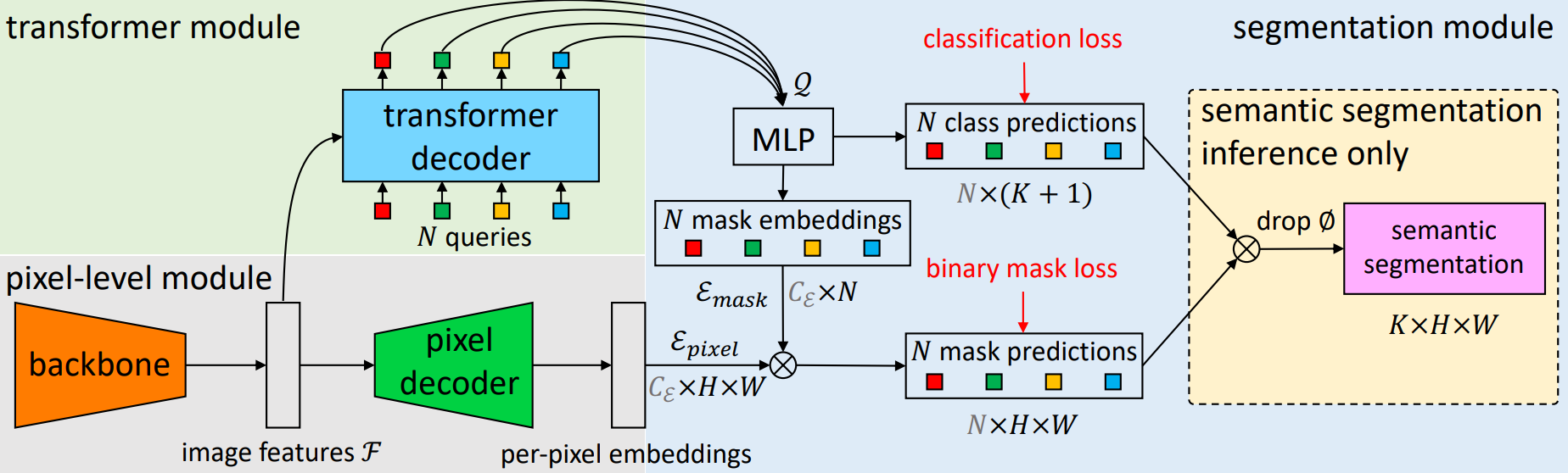}
   \caption{MaskFormer architecture and its three modules: pixel-level, transformer and segmentation. \cite{MaskFormer}}
   \label{fig:maskformer-arch}
\end{figure}

As shown in Figure \ref{fig:maskformer-arch}, the model includes three modules: a pixel-level module to extract embeddings per pixel to generate binary mask predictions, a transformer module to compute per-segment embeddings, and a segmentation module to generate predictions from the embeddings.

\begin{enumerate}
    \item \textbf{Pixel-level module:} This module ingests an $H \times W$ image and using a backbone (Swin Transformer) it generates a low-resolution image feature map $\mathcal{F}\in \mathds{R}^{\mathcal{C}_\mathcal{F}\times \frac{H}{S} \times \frac{W}{S}}$ where $\mathcal{C}_\mathcal{F}$ is the number of channels and $S$ is the stride of the feature map (32). Then, a pixel decoder upsamples the features to generate per-pixel embeddings
    \item \textbf{Transformer module:} Computes per-segment embeddings that encode global information about each segment predicted by MaskFormer. It takes learnable positional embeddings and image features as inputs from the pixel-level module.
    \item \textbf{Segmentation module:} An MLP with 2 hidden layers is used to convert the per-segment embeddings to maks embeddings. Finally, the binary mask prediction per class $m_i\in [0,1]^{H \times W}$ is obtained by computing the dot product between each mask embedding and the corresponding per-pixel embedding from the transformer and per-pixel module,s respectively. The final class probability predictions are computed using a sigmoid activation function on the masks. 
\end{enumerate}

\section{Dataset and features}

\begin{figure}[H]
    \centering
    \begin{adjustwidth}{-\extralength}{0cm}
    \subfloat[\centering]{\includegraphics[width=7.7cm]{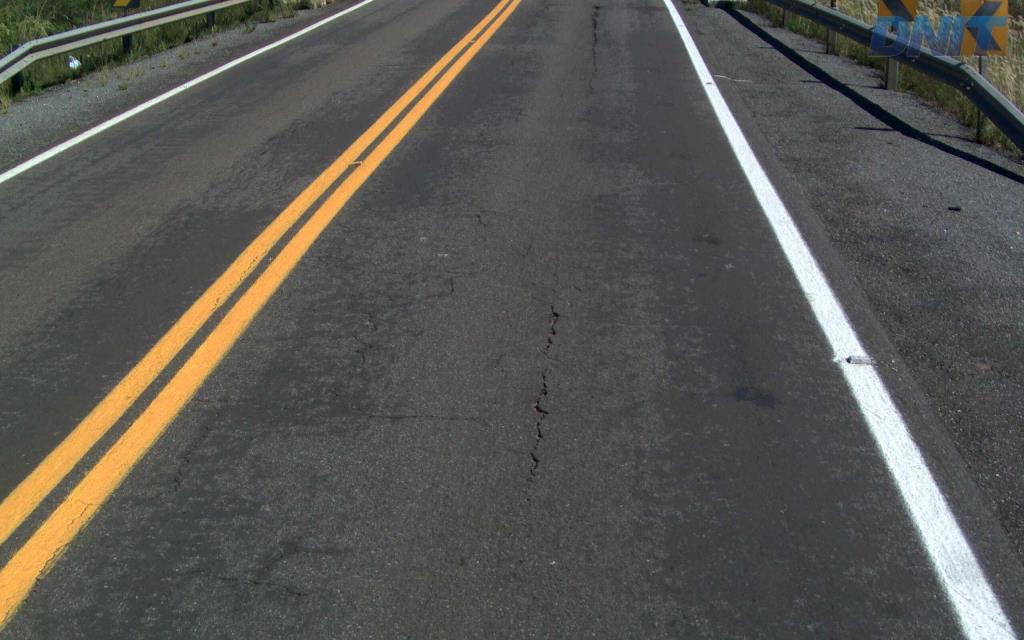}}   
    \hfill
    \subfloat[\centering]{\includegraphics[width=7.7cm]{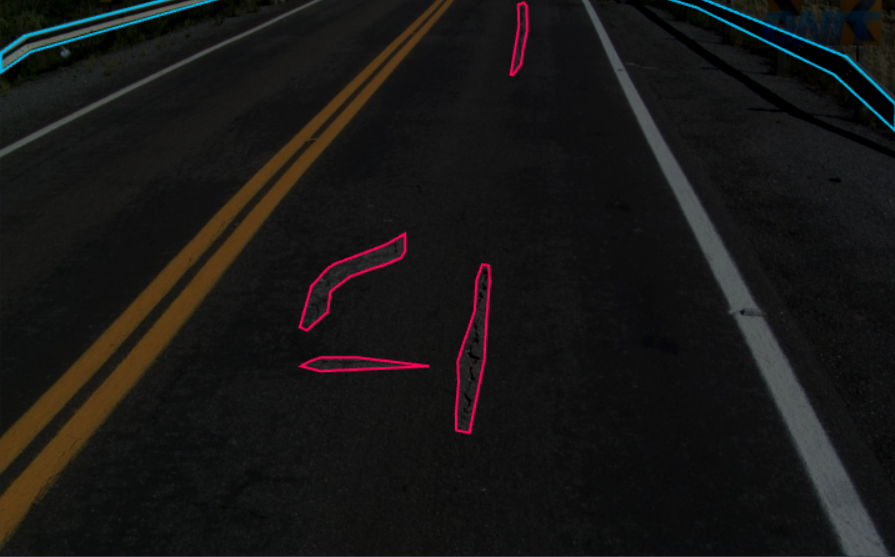}}
    \end{adjustwidth}
    \caption{Example image from the NDTI dataset. (\textbf{a}) Raw image. (\textbf{b}) Annotated images with cracks (red) and damaged guardrail (blue).}
    \label{fig:example}
\end{figure}

The methods described above were tested on the public image dataset provided by the National Department of Transport Infrastructure (NDTI) \cite{brasil_data}, which is available online as an open source with a resolution of $1024 \times 640$ pixels. This dataset includes 2300 images taken in the state of Santa Catarina and has the advantage of being captured in real-world driving conditions, which provides a more realistic and challenging environment for testing the algorithm. However, upon inspection, it was noted that annotations were not complete nor consistent across images. Thus, 1000 of these images were manually annotated, considering 4 different objects of interest: cracks, potholes, damaged guardrails, and damaged lane markings. A sample of the dataset can be observed in Figure \ref{fig:example}. 

\begin{figure}[h]
  \centering
  \includegraphics[width=0.9\linewidth]{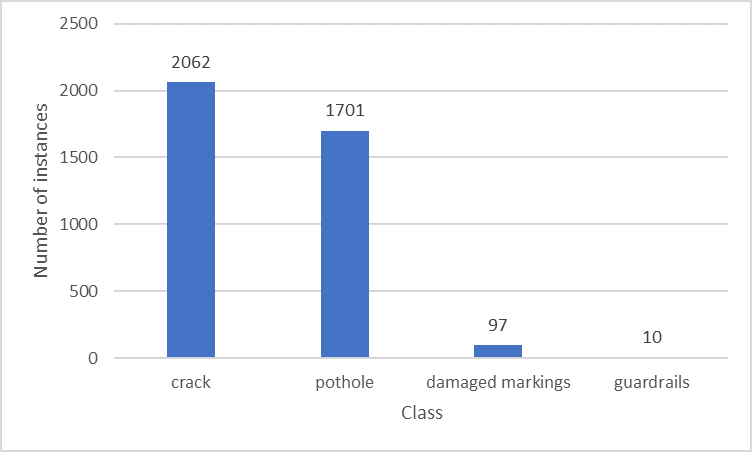}
   \caption{Histogram of number of instances per class in the dataset.}
   \label{fig:hist}
\end{figure}

\begin{figure}[H]
    \centering
    \subfloat[\centering]{\includegraphics[width=5cm]{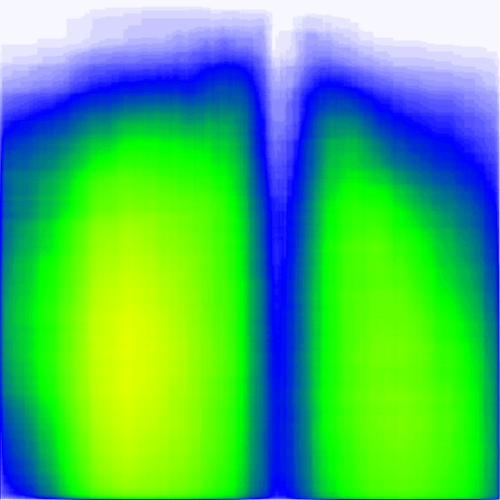}}
    \hfill
    \subfloat[\centering]{\includegraphics[width=5cm]{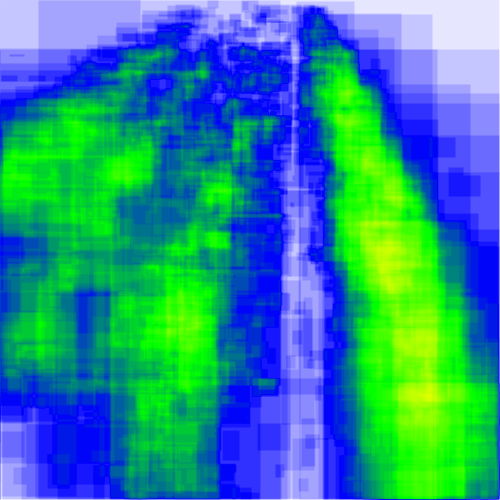}}\\
    \subfloat[\centering]{\includegraphics[width=5cm]{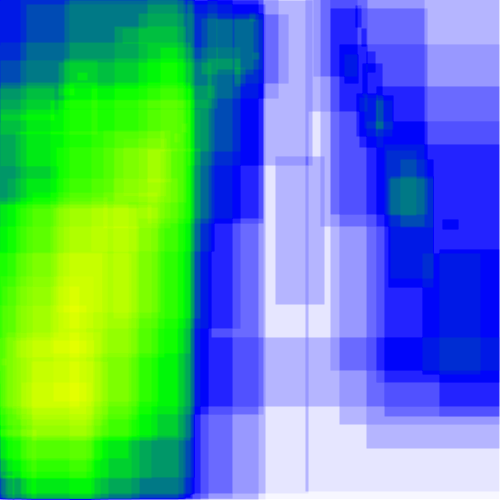}}
    \hfill
    \subfloat[\centering]{\includegraphics[width=5cm]{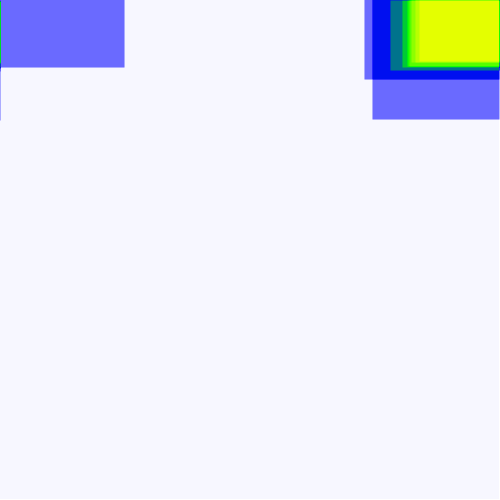}}
    \caption{Annotation heatmaps for each class. (\textbf{a}) Crack. (\textbf{b}) Pothole. (\textbf{c}) Damaged marking. (\textbf{d}) Guardrail.}
    \label{fig:heatmaps}
\end{figure}

The train-valid-test split was set to 85-10-5. The dataset was preprocessed using two operations: auto-orientation and resizing to $640 \times 640$ because the original images caused the virtual machines to consume all of the available RAM. As for augmentations, the following were applied: random crops considering a zoom factor of 20\%, random saturation in the -25\% to 25\% range, and random brightness in the -25\% to 25\% range. One key problem that was observed with the dataset was class imbalance, as the guardrail and lane marking damage categories are severely unrepresented compared to the crack and pothole ones, as seen in Figure \ref{fig:hist}. Further, the annotation heatmaps in Figure \ref{fig:heatmaps}  let us know the location of the features in the images. Cracks and potholes are rarely located in the center of the road, and damaged markings tend to occur mostly in the middle separator (not on the sides).

\section{Results and discussion}
\subsection{Synthetic images from GANs}
 
\begin{figure}[H]
    \centering
    \subfloat[\centering]{\includegraphics[width=0.9\linewidth]{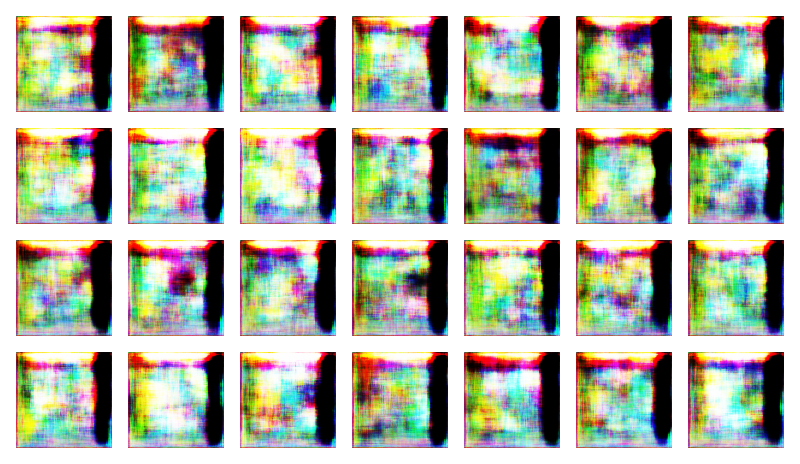}}\\
    \subfloat[\centering]{\includegraphics[width=0.9\linewidth]{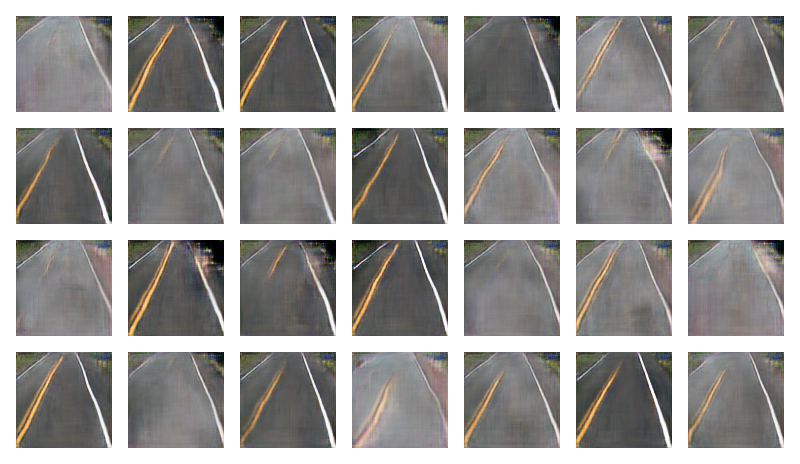}}\\
    \subfloat[\centering]{\includegraphics[width=0.9\linewidth]{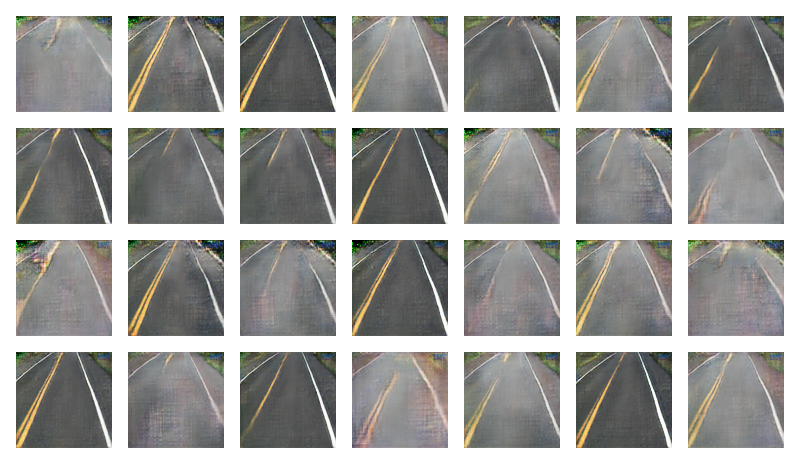}}
    \caption{Synthetic image comparison at different epochs. (\textbf{a}) Epoch=1 (\textbf{b}) Epoch=20 (\textbf{c}) Epoch=50.}
    \label{fig:epochs}
\end{figure}

After training the GAN model for 50 epochs, significant improvements in image quality were observed starting from around epoch 20 (Figure \ref{fig:epochs}). The generated images became more refined and reached a level where they were recognizable as road images in epoch 50. Some images even exhibited details such as cracks and damaged markings, which could potentially be added to the training dataset. The training process demonstrated the capability of the GAN model to generate realistic road images with features resembling real-world road conditions. These results indicate the effectiveness of the model in capturing and reproducing the complex characteristics of road scenes.

\begin{figure}[h]
  \centering
  \includegraphics[width=0.9\linewidth]{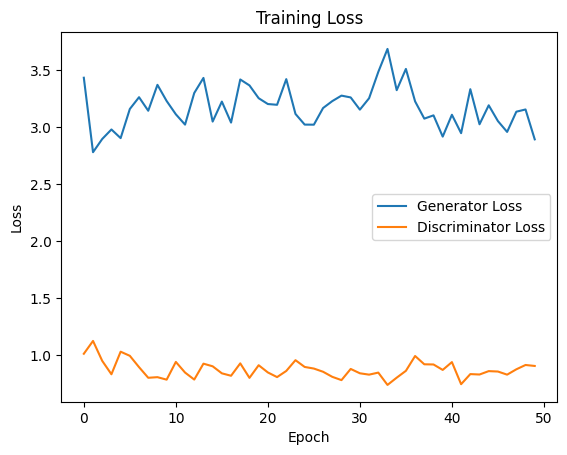}
   \caption{Loss of Generator / Discriminator model during training}
   \label{fig:GAN_Loss}
\end{figure}

During the training process, the losses of the generator and discriminator models were monitored, which served as indicators of the training progress and model performance. The evolution of these values is shown in Figure \ref{fig:GAN_Loss}. The generator loss represents how well this module can produce realistic road images, while the discriminator loss reflects its ability to distinguish between real and generated images. Throughout the training period, a gradual decrease in both losses was observed, indicating the improvement of the generator's performance in synthesizing road images and the discriminator's ability to differentiate between real and artificial images.

Tracking the score of the discriminator is useful to evaluate its accuracy since this module is responsible for differentiating between real and generated samples. Thus, if the generated sample is correctly identified by the Discriminator, it indicates that the generator is producing a more authentic sample. Score trends can be monitored to see if the quality of the generated samples are improving over time.

\subsection{Yolov8 implementation}
\begin{figure}[h]
  \centering
  \includegraphics[width=0.9\linewidth]{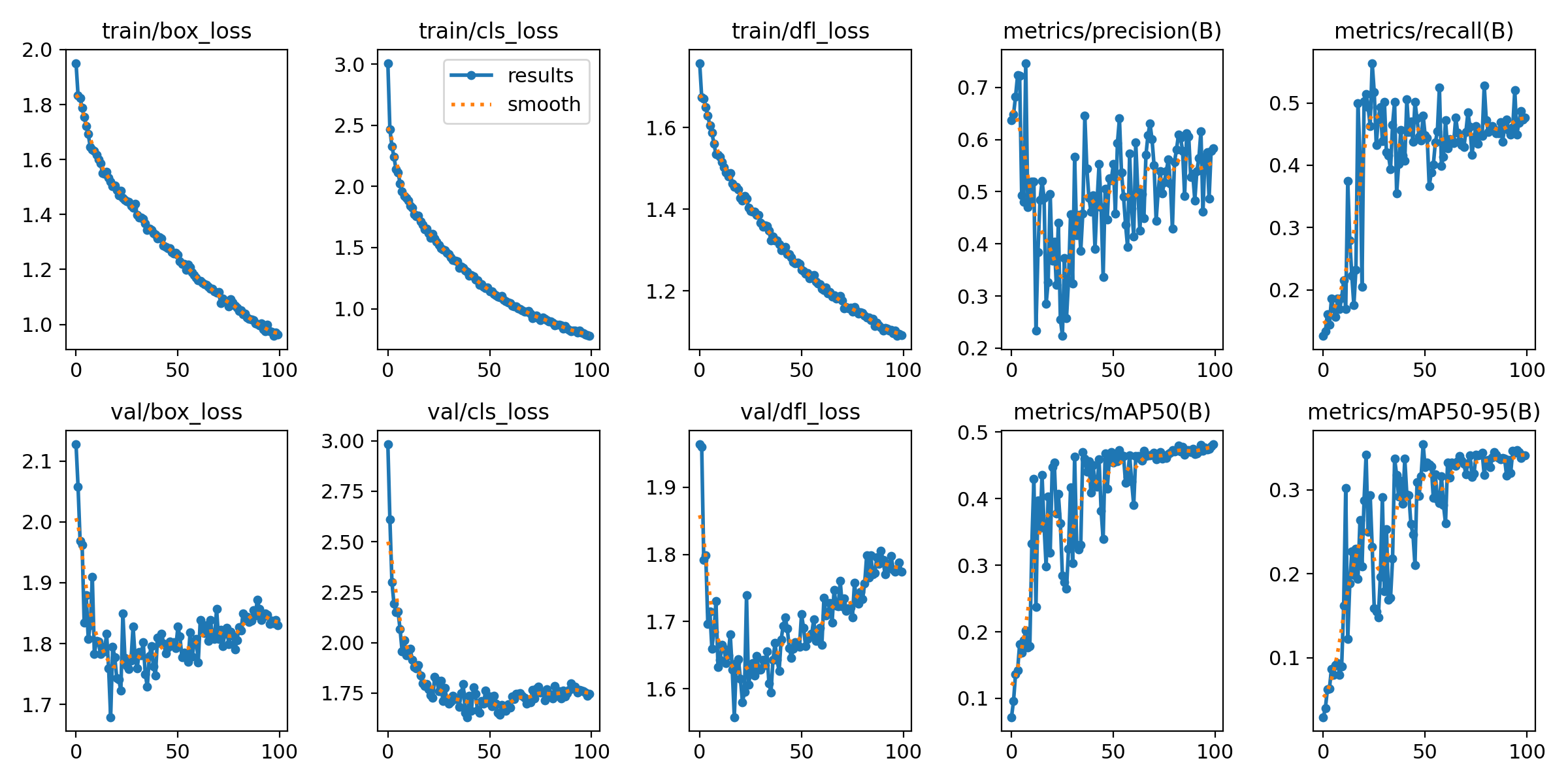}
   \caption{Loss and mAP50 results for YOLOv8 on training (top) and validation (bottom) sets.}
   \label{fig:YOLO_Results}
\end{figure}

 This project used Ultralytics' open-source Git-hub YOLOv8 repository. Hyperparameters were tuned by training for 3 epochs and selecting the highest mAP50. Ultimately, all defaults were kept except the batch size, which was reduced from 16 to 8. The model was then trained for 100 epochs and obtained an mAP50 of 0.471 over all the classes (the highest class being guardrails with mAP50 = 0.995 and the lowest being damaged markings with mAP50 = 0.057). Figure \ref{fig:YOLO_Results} shows the loss (as calculated from the two detection heads) and mAP50 metric computed for each epoch over the training and validation datasets. While the three losses smoothly decrease on the training set, Distribution Focal Loss (DFL) and box loss have a minimum around epoch 25, and the classification loss (CLS) starts plateauing at that time. This indicates the presence of model overfitting to the training data. 
 
\begin{figure}[h]
  \centering
  \includegraphics[width=0.9\linewidth]{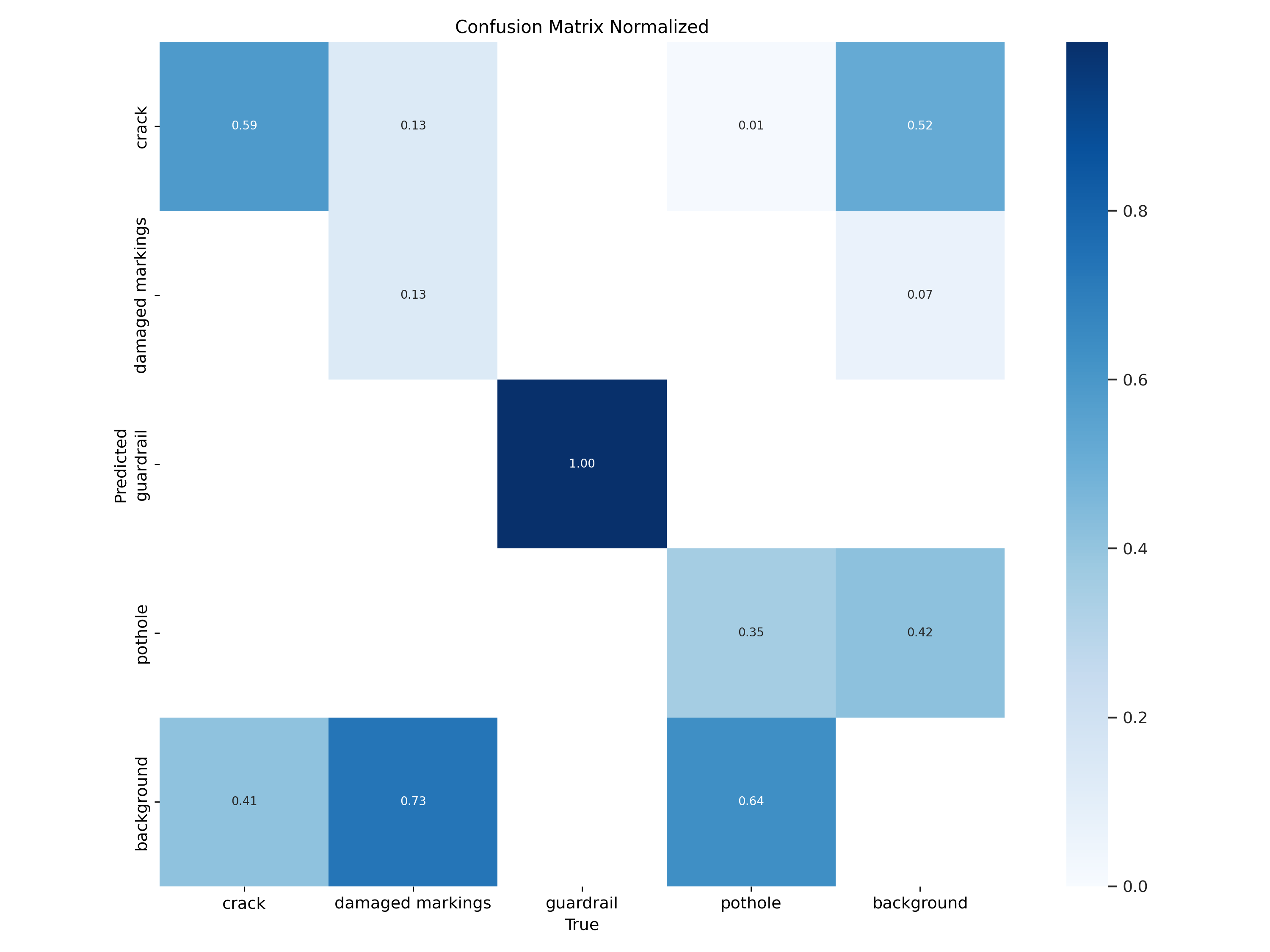}
   \caption{Normalized confusion matrix across classes.}
   \label{fig:Confu_norm}
\end{figure}

The confusion matrix shows that while guardrails are easily identified as such, the three other classes tend to be confused with the background (Figure \ref{fig:Confu_norm}). The class which is most often confused with background is cracks. This was expected since cracks can vary in severity and mild cracks will tend to be confused with healthy concrete, even to the human eye. Additionally, instances of repaired potholes, that might look like potholes but were not labeled as such in the data, as well as old markings that look like damaged markings, might introduce confusion in the model. Labeling more data with more defined classes and precise labels, such as repaired potholes, new marking, old marking, sealed crack, mild crack, medium crack, and severe crack, could help the performance of YOLOv8 in identifying road distress. 
 
\begin{figure}[h]
  \centering
  \includegraphics[width=0.8\linewidth]{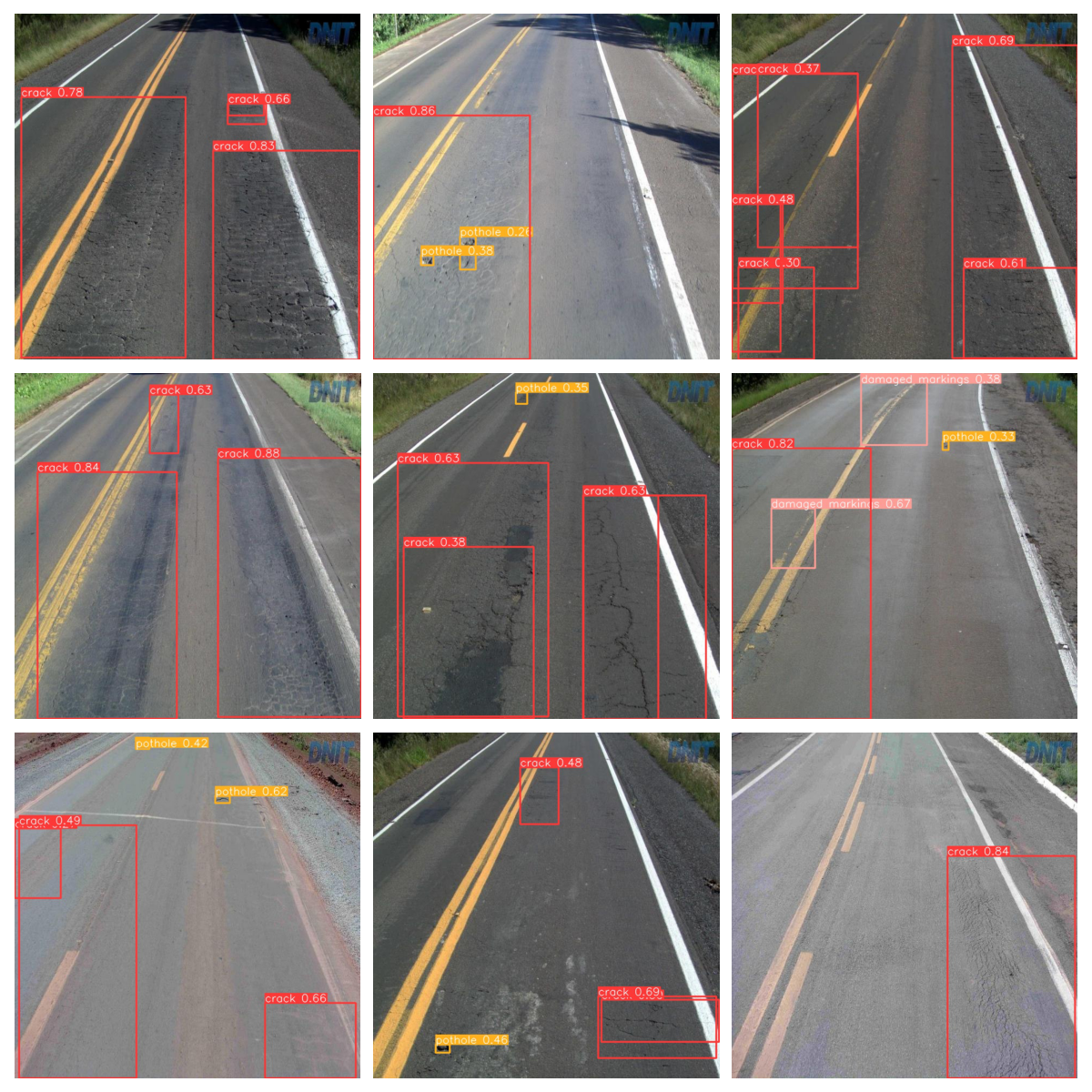}
   \caption{Examples of inference by the YOLOv8 model.}
   \label{fig:YOLO_pred}
\end{figure}

 A random sample of 9 images from the testing set with YOLOv8 predictions is shown in Figure \ref{fig:YOLO_pred}. As it can be seen, YOLOv8 decently predicts most road distresses in these samples, with only one pothole within a crack in the center image not being identified. 

\subsection{MaskFormer implementation}
\begin{figure}[H]
  \centering
  \includegraphics[width=0.8\linewidth]{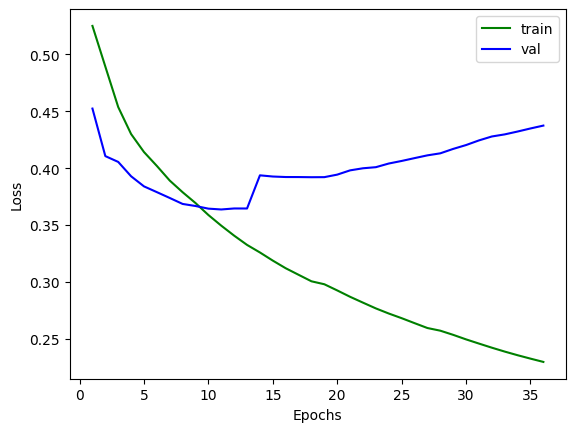}
   \caption{Loss evolution during training and validation for the Maskformer model.}
   \label{fig:loss_trans}
\end{figure}

\begin{figure}[H]
  \centering
  \includegraphics[width=0.8\linewidth]{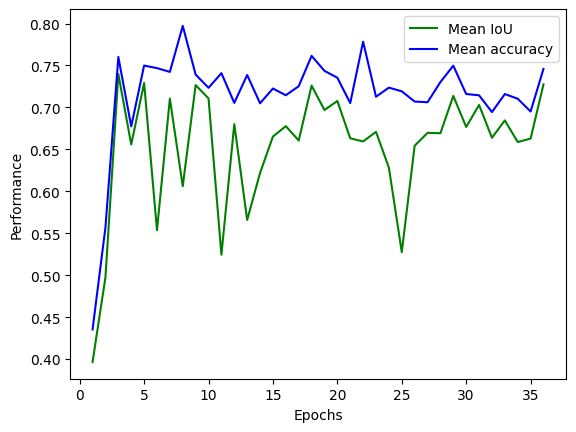}
   \caption{IoU and accuracy evolution during validation for the Maskformer model.}
   \label{fig:IoU_trans}
\end{figure}

The best transformer model for segmentation was obtained using an Adam optimizer with $\beta_1$ and $\beta_2$ values of 0.9 and 0.999, respectively, as well as a learning rate of 0.00005. The model was trained for 35 epochs and obtained the following metrics on validation: 0.727 for mean IoU, and 0.746 for mean accuracy. In Figures \ref{fig:loss_trans} and \ref{fig:IoU_trans}, the evolution of loss values and IoU/accuracy values can be observed. From these charts, it can be seen that the validation loss was minimized around epoch 15, and then the model started to overfit as the training loss kept decreasing, but the validation loss started increasing. Using a best model selector script developed by the researchers, this model was selected for testing. As for the performance metrics, it can be observed that the behavior of accuracy is far more stable compared to IoU. Further, IoU and accuracy are not maximized at the point where loss is minimized on the validation dataset. This could be because, although the model with the lowest loss minimizes the probability of wrong segmentations, it still does not output the correct category, thereby displaying subpar values for IoU and accuracy. 

\begin{figure}[h]
    \centering
    \begin{adjustwidth}{-\extralength}{0cm}
    \subfloat[\centering]{\includegraphics[width=7.7cm]{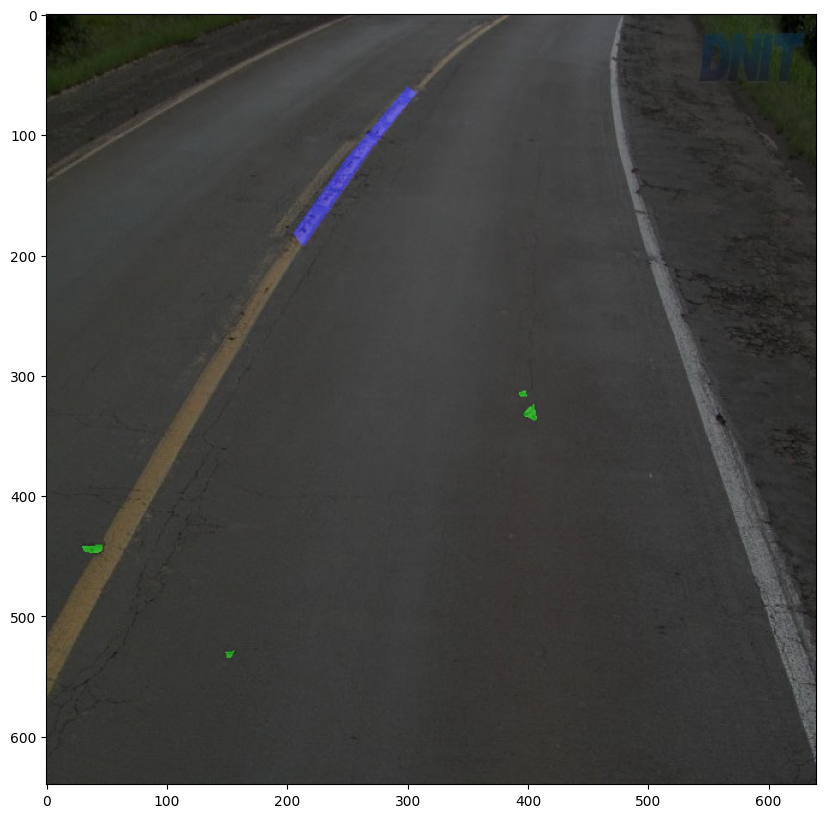}}   
    \hfill
    \subfloat[\centering]{\includegraphics[width=7.7cm]{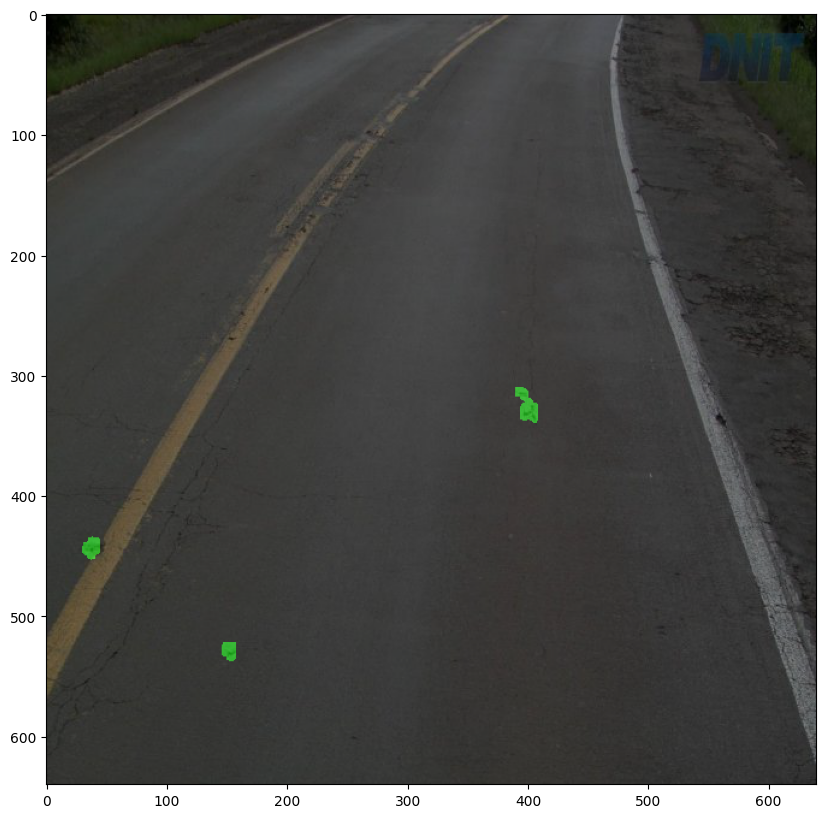}}
    \end{adjustwidth}
    \caption{Example of segmentation results using MaskFormer on the testing dataset. (\textbf{a}) Ground truth. (\textbf{b}) Annotations by MaskFormer.}
    \label{fig:example_maskformer}
\end{figure}

On the testing dataset, the following results were obtained: 0.707 for mean IoU, and 0.723 for mean accuracy. Examples of the segmentation results during testing are presented in Figure \ref{fig:example_maskformer}. Qualitatively it can be seen that the transformer model has optimal performance in the class related to cracks and potholes. As these two categories are greatly represented in the dataset, the model was able to generalize adequately and learn from their features. Conversely, the damaged markings and guardrails, which are severely underrepresented, were not segmented correctly on the test dataset, if at all. Most likely, the data was insufficient to yield any significant learning results for the transformer. Analyzing the logits present in the masking outputs, however, showed that the logits are not null. In the context of Figure \ref{fig:example_maskformer}, this means that the model has some intuition of the presence of lane marking damage in the dataset but lacks enough certainty to draw the mask and opts to leave it unlabeled.

\section{Conclusions}

In the first step, we successfully utilized GANs to generate new images containing roads, cracks, potholes, and other features. However, the generated images still lack the desired level of detail required for practical training purposes. While the GAN framework showed promise in generating diverse samples, the implementation of higher resolution images over 640×480 is necessary to make the generated images more visually convincing and suitable for use as training data. Also, at this stage, the generated images need to be manually annotated. Future development could be to introduce semi-supervised learning, which would allow for more efficient automated annotation.

Subsequently, the implementation of YOLOv8 achieved a mean average precision of 0.471 when a bounding box is considered a true positive if it has an IOU of 50\% or more with the ground truth. This means that the model is relatively good at detecting objects, but there is room for improvement. The performance of YOLOv8 for road distress detection could be improved through further hyperparameter tuning (reducing the learning rate could help prevent over-fitting), by using a larger dataset with a better quality of annotations and by applying custom data augmentations that focus on the distresses of interest. 

The MaskFormer implementation showed promising results with an IoU of 0.707 and accuracy  of 0.723 for the test dataset. The model excelled at the segmentation of cracks and potholes, but were severely lacking in the underrepresented classes: damaged markings and guardrail damage. For future work, it is imperative to balance the dataset including more samples of the underrepresented classes. Further, with more compute resources available, it is recommended to explore more complex models such as OneFormer \cite{jain2023oneformer}, a powerful panoptic segmentation model that was released during the development of this project.

As the threshold in the IoU was set to 0.50, the mAP50 and IoU results for Yolov8 and MaskFormer are comparable. Quantitatively, MaskFormer yields overall the best results, which is expected given the model's complexity. However, qualitative analysis results on the test set reveal that Yolov8 appears to identify more objects of interest from underrepresented classes compared to the transformer model. 

\authorcontributions{All authors have contributed equally to conceptualization, methodology, software, validation, formal analysis, investigation, resources, data curation, and writing. All authors have read and agreed to the published version of the manuscript.}

\funding{This research received no external funding}

\dataavailability{The data presented in this study are available in Mendeley Data at https://data.mendeley.com/datasets/t576ydh9v8/4. These data are directly derived from the above resource, which is available in the public domain.} 

\acknowledgments{We would like to express our deepest gratitude to Professor Fei-Fei Li for her invaluable guidance and mentorship throughout this project. Her insights and expertise in computer vision were instrumental in shaping the direction and depth of this work. I am especially grateful for her thoughtful feedback and continuous encouragement, which have been a source of inspiration throughout this research.}

\conflictsofinterest{The authors declare no conflicts of interest.} 

\begin{adjustwidth}{-\extralength}{0cm}

\reftitle{References}


\bibliography{bibliography}

\PublishersNote{}
\end{adjustwidth}
\end{document}